\documentclass[10pt,twocolumn,letterpaper]{article}

\usepackage{cvpr}
\usepackage{times}
\usepackage{epsfig}
\usepackage{graphicx}
\usepackage{amsmath}
\usepackage{amssymb}

\usepackage[pagebackref=true,breaklinks=true,letterpaper=true,colorlinks,bookmarks=false]{hyperref}

\cvprfinalcopy 

\def\cvprPaperID{34} 

\ifcvprfinal\fi
\begin{document}



\title{Multimodal Age and Gender Classification Using Ear and Profile Face Images}

\author{Dogucan Yaman\textsuperscript{*} \hspace{0.8cm} Fevziye Irem Eyiokur\textsuperscript{*} \hspace{0.8cm} Haz{\i}m Kemal Ekenel \\
Istanbul Technical University, Turkey\\
{\tt\small \{yamand16, eyiokur16, ekenel\}@itu.edu.tr }
}

\maketitle
\thispagestyle{empty}
\renewcommand{\thefootnote}{\Alph{footnote}}
\footnotetext{\textsuperscript{*}The authors have equally contributed.}

\begin{abstract}
   In this paper, we present multimodal deep neural network frameworks for age and gender classification, which take input a profile face image as well as an ear image. Our main objective is to enhance the accuracy of soft biometric trait extraction from profile face images by additionally utilizing a promising biometric modality: ear appearance. For this purpose, we provided end-to-end multimodal deep learning frameworks. We explored different multimodal strategies by employing data, feature, and score level fusion. 
   To increase representation and discrimination capability of the deep neural networks, we benefited from domain adaptation and employed center loss besides softmax loss. We conducted extensive experiments on the UND-F, UND-J2, and FERET datasets. Experimental results indicated that profile face images contain a rich source of information for age and gender classification. We found that the presented multimodal system achieves very high age and gender classification accuracies. Moreover, we attained superior results compared to the state-of-the-art profile face image or ear image-based age and gender classification methods.
\end{abstract}

\section{Introduction}\label{Section Introduction}

Estimating soft biometric traits is a popular research area in biometrics \cite{jain2004, jain2009, vaquero2009, saeed2018}. It has been shown that the soft biometric traits enable to describe subjects better and affect the identification performance affirmatively \cite{jain2004, jain2009, ozbulak2016transferable}. Age and gender are among the mostly used soft biometric traits. 

\begin{figure}
\begin{center}
\includegraphics[width=0.8\linewidth]{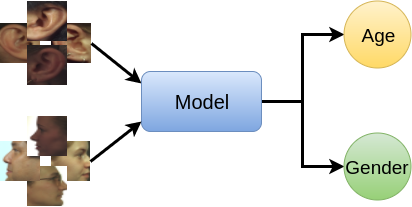}
\end{center}
   \caption{Overview of the multimodal, multitask age and gender classification framework. }
\label{fig:problem_definition}
\end{figure}

In recent years, deep convolutional neural network (CNN) based approaches \cite{levi2015age, rothe2018deep, yaman2018age} have been commonly used for automatic age and gender classification. Generally, frontal or close to frontal face images have been used in these studies \cite{levi2015age, ozbulak2016transferable, he2017deep, zhang2018fine, rothe2018deep}. There have also been studies about gender classification from ear images \cite{gnanasivam2013gender, khorsandi2013gender, lei2013gender,yaman2018age, iannarelli1989ear, abaza2013survey, pflug2012ear, sforza2009age, purkait2007anthropometry, more_than_ziga} and profile face images~\cite{bukar2017automatic}, and combination of them as well \cite{zhang2011hierarchical}. However, there is only one recent work \cite{yaman2018age} that performs age classification from ear images and one that performs age estimation from profile face images \cite{bukar2017automatic}.

In this paper, we presented various end-to-end multimodal deep CNN frameworks, which performed multitask learning for age and gender classification using ear and profile face images as shown in Figure \ref{fig:problem_definition}. 
Although profile face images, when cropped using a large bounding box, can contain ear appearance, the main motivation to utilize a multimodal system instead is 
to avoid the presence of irrelevant features. That is, when we expand the bounding box of the profile face images, hair and background information are also included in the image, which we found to degrade the performance in our experiments. 
In the proposed multimodal networks, we explored three different fusion methods. These are: Data fusion ---intensity fusion, spatial fusion, channel fusion---, feature fusion, and score-level fusion. As deep CNN models, we utilized VGG-16 \cite{VGG} and ResNet-50 \cite{resnet}. To increase representation and discrimination capability of the multimodal deep neural networks, we benefited from domain adaptation and employed center loss besides softmax loss. In summary, our contributions are listed as following:
\begin{itemize}
    \item We presented multimodal age and gender classification systems that take profile face and ear images as input. The proposed systems perform end-to-end multimodal, multitask learning.
    \item We have comprehensively explored various ways of utilizing multimodal input for age and gender classification. We employed three different data fusion methods, as well as feature and score level fusion.
    \item We performed domain adaptation in order to adapt the pretrained CNN models, namely VGG-16 and ResNet-50, to the ear domain. For this, we generated the extended version of the Multi-PIE ear dataset that was presented in a previous work \cite{our_ear_journal} and named it Multi-PIE extended-ear dataset. Moreover, to learn more discriminative features, we utilized center loss in combination with the softmax loss. 
    \item  We conducted extensive experiments on the UND-F, UND-J2, and FERET datasets for gender classification, and only on the FERET dataset for age classification, since no age information is available in the UND-F and UND-J2 datasets. 
    We achieved state-of-the-art age and gender classification results on these datasets. 
    \item We presented the first study on age classification using ear images that is conducted on a publicly available dataset, namely, FERET, in contrast to the previous work \cite{yaman2018age}, in which an internal dataset was used. In addition, compared to this previous work, we improved the age classification accuracy from 52\% to 67.59\% for classifying five age groups.

\end{itemize}

\section{Related Work}\label{Section Related Work}
In this section, we briefly reviewed the age and gender classification studies that use ear and profile face images.

In \cite{gnanasivam2013gender}, authors used ear-hole as a reference point and the distance between this point and additional seven points are calculated to extract features. With these features, the best result is achieved by a k-nearest neighbor classifier with 90.42\% accuracy on an internal dataset, which contains 342 samples. In \cite{zhang2011hierarchical}, ear and profile face images are used for the gender classification. Support vector machine (SVM) classifier is employed with histogram intersection kernel. Score level fusion is utilized to increase the performance. Experiments are conducted on the 2D ear images of the UND-F dataset. Multimodal system's performance is found to be 97.65\%, while face-only accuracy is 95.42\% and ear-only accuracy is 91.78\%. In \cite{khorsandi2013gender}, features are extracted with Gabor filters and these features are then classified using majority voting. 
In total, 128-dimensional features are utilized and 89.49\% classification accuracy is achieved on the UND-J2 dataset. In \cite{lei2013gender}, gender classification experiments are conducted on both 2D and 3D ear images of UND-F and UND-J2 datasets. In the experiments, Histogram of Indexed Shapes features are extracted, SVM is used for classification and 92.94\% accuracy is obtained. In \cite{bukar2017automatic}, authors investigated the usability of profile face image in such cases that frontal face cannot be captured. The experimental results are conducted on four different datasets. Three types of ResNet architectures, ResNet-50, ResNet-101, ResNet-152, are employed for feature extraction and sparse partial least-squares regression is used with extracted features. According to the experiments on FERET dataset, the best result is obtained using ResNet-152 features with 5.50 mean absolute error (MAE). In \cite{yaman2018age}, the authors present a study on age and gender classification from ear images. They employed both a geometric-based ---distances between ear landmarks and area information--- and an appearance-based representation ---deep CNNs. The authors conducted their experiments on an internal dataset and found that appearance-based representation is more useful \cite{yaman2018age}. 

In this work, we improved the previous results on age and gender classification from ear and profile face images by benefiting from domain adaptation, center loss, and multimodal fusion. We have proposed end-to-end multimodal deep learning frameworks and further increased the classification accuracies.

\section{Methodology}\label{Section Methodology}
In this section, we present employed CNN models and explain proposed multimodal, multitask approach. We also describe the benefited transfer learning and domain adaptation strategies.

\subsection{CNN Models and Loss Functions}\label{Section CNN}
We employed two different well-known deep CNN models, which are VGG-16 \cite{VGG} and ResNet-50 \cite{resnet}. In VGG-16, there are 13 convolutional layers and 3 fully-connected layers. To prevent overfitting, dropout method~\cite{dropout} was employed. Softmax loss was used in order to produce probabilities for the classification task. In addition, in this work we utilized combination of softmax loss and center loss. We also used weight decay as a regularizer in the training phase in order to prevent overfitting. Weight decay parameter is set to 0.001 in the experiments. The other deep CNN model that is used in this work is ResNet-50. 
In contrast to the VGG-16, there are no fully-connected layers except the output layer in the ResNet-50. There exists a global pooling layer between the convolutional part and the output layer. The input size of both of these networks is $ 224 \times 224 $.

We utilized center loss \cite{center_loss} with softmax loss in order to obtain more discriminative features. The main motivation behind center loss is to provide features that are closer to corresponding class center. The distance between features and related class center is measured and the total center loss is calculated. Then, center loss part and softmax loss are summed to calculate the final loss. The center loss tries to produce closer features for each class center but it is not responsible of providing separable features. Thus, this is complemented by softmax loss. Besides, for the total loss value, there is a coefficient for center loss that determines the effect over total loss. The formula of the total loss is presented in Equation \ref{eq_softmax_and_center_loss}. The first part of the formula is the softmax loss and the second part is the center loss. In the center loss formula, $ c_{y_i} $ represents $y_i$th class center and $ x_i $th feature. In the experiments, $\lambda$ coefficient is selected as 0.1 according to the results obtained on the validation set.

\begin{equation}
\label{eq_softmax_and_center_loss}
   L = - \sum_{i=1}^{m} log \frac{e^{W^T_{y_i}x_i + b_{y_i}}}{\sum_{j=1}^{n} e^{W^T_j x_i + b_j}} + \frac{\lambda}{2} \sum_{i=1}^{m} ||x_i - c_{y_i}||_2^2 
\end{equation}

\begin{figure*}
\begin{center}
\includegraphics[width=0.95\linewidth]{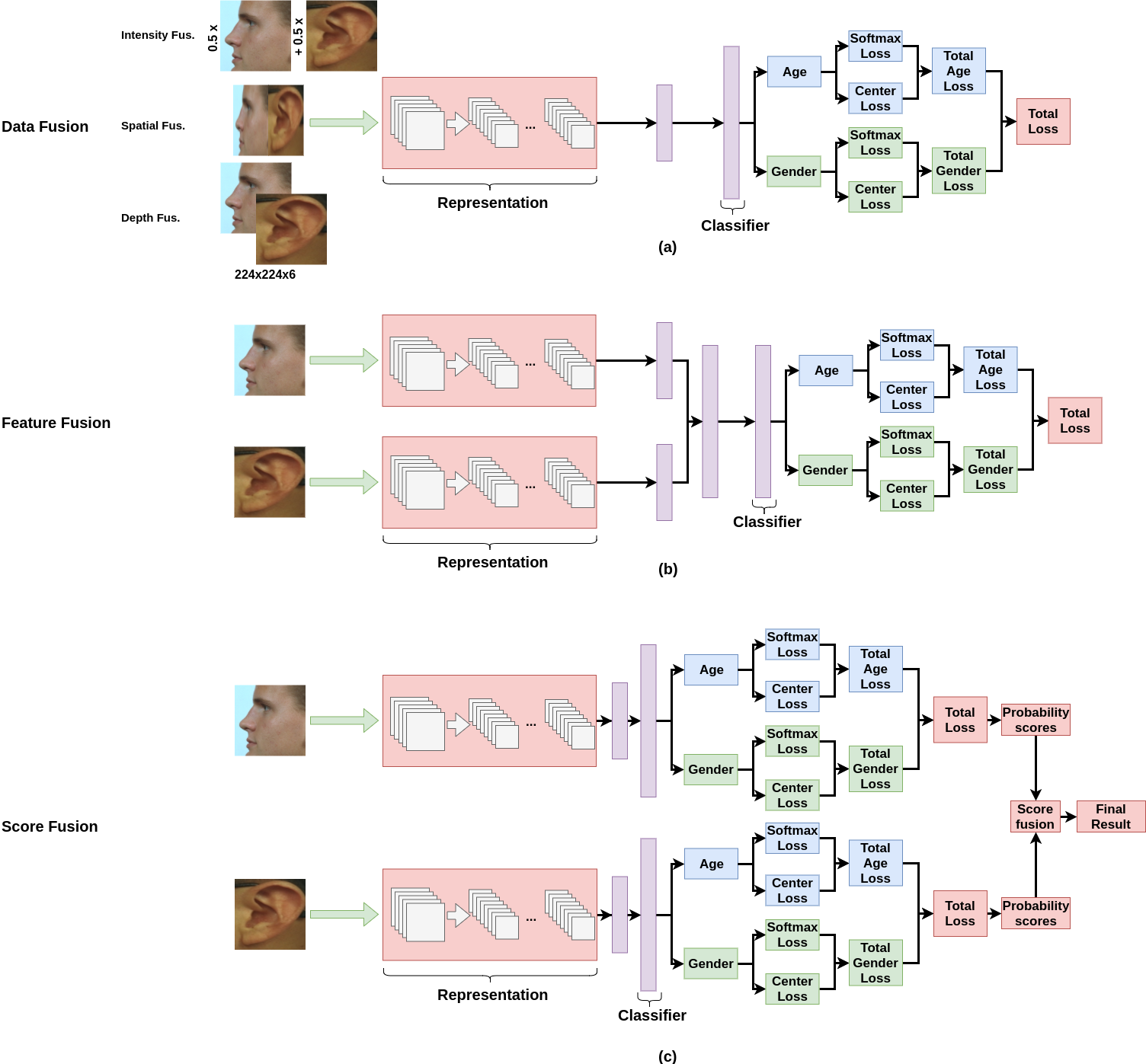}
\end{center}
   \caption{Multimodal fusion methods. (a) presents employed three different data fusion methods. In the first one, named as intensity fusion, we have averaged profile face and ear images' pixel intensity values. In the second approach, spatial fusion, we have combined profile face and ear images side-by-side. In the last method, channel fusion, we have concatenated profile face and ear images in depth, i.e. along their channels. (b) presents feature fusion. The representation part of the networks has been trained separately and outputs of these parts have been concatenated. This concatenated feature vector then fed to the classification part. In (c), score fusion, two different networks have been trained separately, then score-based concatenation has been performed according to the probability scores of the networks. 
   }
\label{fig:fusion_methods}
\end{figure*}

\subsection{Multimodal and Multitask CNN}\label{Section Multimodal and Multitask Convolutional Neural Network}

We investigated the performance of age and gender classification using ear and profile face images both separately, as unimodal systems, and together as a multimodal, multitask system. As can be seen from Figure \ref{fig:fusion_methods}, we explored three end-to-end multimodal architectures. 
For all experiments, we have employed VGG-16 \cite{VGG} and ResNet-50 \cite{resnet} models with center loss \cite{center_loss}. For the total loss calculation in multimodal, multitask age and gender classification experiments, we have combined all losses that come from age and gender prediction of the network. The final loss is calculated using the formula presented in Equation \ref{eq_total_loss_calculation}.

\begin{equation}
\label{eq_total_loss_calculation}
    L_{total} = \beta * ( L_{{total}_{age}} ) + ( 1 - \beta ) * ( L_{{total}_{gender}} )
\end{equation}

According to the above formula, the calculation of total age and gender losses are based on summation of the softmax loss and multiplication of the center loss with $ \lambda $ coefficient as explained before. After the measurements of the total age and gender losses, we have combined these loss values using $ \beta $ coefficient. The main motivation behind introducing the $ \beta $ coefficient is that since gender classification result is significantly high, we have tried to highlight age classification loss for the final loss. With this coefficient, we can basically change the effect of age-specific and gender-specific losses over total loss. 
In the experiments, we have tried several different $ \beta $ coefficient and according to the obtained results on the validation set, we have achieved the best performance with $ \beta = 0.75 $ value.

\subsection{Fusion Methods}\label{Section Fusion Methods}
In this section, we present the multimodal fusion methods that are illustrated in Figure \ref{fig:fusion_methods}. 

\begin{figure}
\begin{center}
\includegraphics[width=0.8\linewidth]{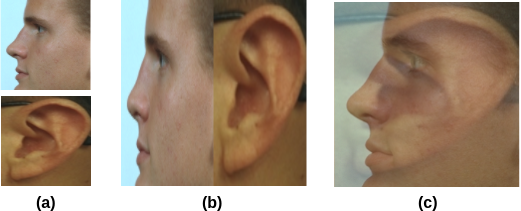}
\end{center}
   \caption{Visualization of employed data fusion approaches. (a) contains original ear and profile face images. (b) represents side-by-side concatenation, which is spatial fusion and (c) is the average of the profile face and ear images, that is intensity fusion. 
   }
\label{fig:image-based_fusion}
\end{figure}

\subsubsection{Data Fusion}\label{Section Image-based Fusion}
To perform data fusion, we have employed three different methods, namely spatial fusion, intensity fusion, and channel fusion. In spatial fusion, we concatenate profile face and ear images side-by-side. 
In the concatenated image, left half of the image contains profile face image and right half includes ear image, as can be seen from Figure \ref{fig:image-based_fusion} (b). 
In channel fusion, we have concatenated images along channels. That is, our input data is [224, 224, 3] dimensional, after channel-based concatenation our input data becomes [224, 224, 6]. While the first three channels belong to profile face image, the remaining 3 channels contain ear image. Finally, in intensity fusion, we average pixel intensity values of the profile face and ear images as shown in the Figure~\ref{fig:image-based_fusion} (c). 


\subsubsection{Feature Fusion}\label{Section Feature-based Fusion}
For the feature-based fusion strategy, we have trained two separate CNN models, one of them takes profile face image as input and the other one takes ear image as input. While the representation part (convolutional part) of these networks have been trained separately, the outputs of the last convolution layers have been concatenated and fed to the classifier part. For example, for VGG-16, thirteen convolutional layers are the separate part of the networks and three fully connected layers are the common part of the multimodal system. In VGG-16, the size of the output vector of each representation is $ 1 \times 4096 $. Two vectors are then concatenated making a $ 1 \times 8192 $ dimensional feature vector and fed to the fully connected layers. 
For ResNet-50, we have combined the outputs of the last layer before global average pooling layer, then the combined vector is passed through global average pooling layer and a fully-connected layer, respectively. 

\subsubsection{Score Fusion}\label{Section Score-based Fusion}
In this method, we have performed two individual training with different CNN models. While one CNN model has been trained on profile face images, the other one has been trained with ear images. Then, for the score-based fusion, we have tested each profile face image and ear image with related models. After that, for each profile face and ear image that belong to the same subject, we have acquired probability scores and measured the confidence score of each model according to five different confidence score calculation methods that are presented in Table \ref{table:score-based_fusion_formulas}. Later, we have selected the prediction of the model that has the maximum confidence score. 

\subsection{Transfer Learning and Domain Adaptation 
}
\label{Section Transfer Learning and Domain Adaptation}

For convolutional neural networks, it is very common and have been shown to be useful \cite{yosinski2014transferable, sharif2014cnn, ozbulak2016transferable} to benefit from previously trained models, more specifically the models that have been trained on the large-scale ImageNet dataset \cite{imagenet_dataset}. This approach, transfer learning, helps to adapt the successful CNN models to the problems, where only a limited amount of data is available. A way of applying transfer learning is to perform fine-tuning. In this approach, the parameters of the pretrained model are used to initialize CNN models' weights instead of using random initial values. Then these model weights are further updated using the target dataset. During fine-tuning, depending the similarity between target dataset domain and pretrained dataset domain, and the size of the target dataset, some layers' weights can be frozen or all layers' weights can be updated. 

\begin{table}
\begin{center}
\begin{tabular}{c|c}
\hline
Method & Formula \\
\hline\hline
Basic & $c = s[0]$ \\
d2s & $c = s[0] - s[1]$ \\
d2sr & $c = 1 - (s[1]/s[0])$ \\
avg-diff & $c = \frac{1}{M-1} \sum_{i=1}^M (s[0] - s[i])$ \\
diff1 & $c = \sum_{i=1}^{M-1} (\frac{s[i-1] - s[i]}{i})$ \\
\hline
\end{tabular}
\end{center}
\caption{Confidence score calculation methods for score-based fusion. In the formulas, c represents confidence score of the corresponding CNN model and s is an array that contains probabilities for each class from high to low value. 
}
\label{table:score-based_fusion_formulas}
\end{table}

Since it has been shown that \cite{ozbulak2016transferable} transferring a pretrained model from a closer domain leads to better performance, we have benefited from a large-scale ear dataset, which is named as Multi-PIE ear dataset \cite{our_ear_journal}. As the name implies, this dataset was constructed by running an automatic ear detector on the profile and close-to-profile face images of the Multi-PIE face dataset \cite{multi_pie}. This way, 17183 ear images of 205 subjects were detected \cite{our_ear_journal}. In this work, we have extended Multi-PIE ear dataset \cite{our_ear_journal} and named it as Multi-PIE extended-ear dataset. The coordinates of the detected ear images of the Multi-PIE ear dataset \cite{our_ear_journal} have been shared \footnote{https://github.com/iremeyiokur/multipie\_ear\_dataset}. We have used these ear coordinates 
and additionally, we have executed our improved ear detection algorithm on the other images of Multi-PIE dataset that were not listed in the Multi-PIE ear dataset. This way, we have acquired new ear images in addition to the ones in the Multi-PIE ear dataset \cite{our_ear_journal} and reached 39185 ear images. The coordinates of this extended-ear dataset will be shared \footnote{https://github.com/iremeyiokur/multipie\_extended\_ear\_dataset} as well. Compared to the Multi-PIE ear dataset, we provided a significant increase in the dataset size. For training, we have first initialized the CNN models with the parameters of the pretrained CNN models that were trained on the ImageNet dataset. Then, by utilizing the Multi-PIE extended-ear dataset, we have updated the CNN models and adapt them to the ear domain. Afterwards, we have benefited from adapted pretrained models and further updated them by training on the target ear datasets. 
Experimental results have validated that this learning strategy, that is by performing domain adaptation via an intermediate fine-tuning step, and by this way using the CNN models that were adapted to the ear domain instead of directly using the CNN models that were trained on a generic image classification dataset, helped to improve the performance of both age and gender classification tasks.

\section{Experimental Results}\label{Section Experimental Results}
In this section, we provide information about used datasets, experimental setups, implementation details, and our results, respectively. 

\subsection{Datasets}\label{Section Datasets}

\textbf{Multi-PIE Extended-ear Dataset} contains 39185 ear images. 
This dataset was constructed by running an automatic ear detector on the profile and close-to-profile face images of the Multi-PIE face dataset~\cite{multi_pie}. 
As explained in the Section \ref{Section Transfer Learning and Domain Adaptation}, this dataset was used to adapt the pretrained CNN models to the ear domain.

\begin{figure}
\begin{center}
\includegraphics[width=0.8\linewidth]{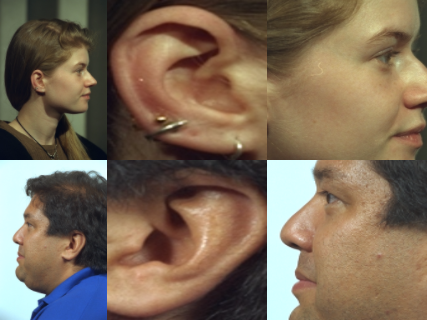}
\end{center}
   \caption{Sample images from FERET dataset \cite{FERET_dataset}. 
   The first column contains the original images, the second column contains sample detected ear images, and the last column contains detected profile face images.}
\label{fig:feret_dataset}
\end{figure}

\textbf{FERET \cite{FERET_dataset}} is one of the most well-known face datasets. Sample images from FERET are shown in Figure~\ref{fig:feret_dataset}. For our experiments, we used both ear and profile face images. We utilized dlib library \cite{dlib} to detect profile faces. In order to detect the ear regions, we run an off-the-shelf ear detector \cite{opencv}. This way we obtained 1397 ear images. 


\textbf{UND-F \cite{UND_datasets}} contains both 2D and 3D ear images. In this work, we only used 2D ear images. There are 942 profile face images that belong to 302 different subjects. We run the aforementioned off-the-shelf ear detector \cite{opencv} and crop ear regions from profile face images. We utilized dlib library \cite{dlib} for detecting profile face images. This dataset was employed to benchmark gender classification accuracy, since it was also used in the previous work on gender classification from ear images \cite{lei2013gender, zhang2011hierarchical}.

\textbf{UND-J2 \cite{UND_datasets}} is another dataset that was utilized to benchmark gender classification accuracy in previous work \cite{khorsandi2013gender, lei2013gender}. It includes 2430 profile face images. As in the UND-F dataset, we detected the ear regions by applying the off-the-shelf ear detector \cite{opencv} and we employed dlib \cite{dlib} to detect profile faces. 

\subsection{Implementation Details}\label{Section Implementation Details}
In order to have domain adapted pretrained deep models, we have performed fine-tuning on Multi-PIE extended ear dataset with VGG-16 \cite{VGG} and ResNet-50 \cite{resnet} CNN models. For this, we have splitted Multi-PIE extended ear dataset into training, validation, and test sets. 80\% of the images have been assigned to the training set, and the remaining 20\% has been assigned evenly to the validation and test sets. The same percentage of training, validation, and test splits are also applied in the age and gender classification experiments on FERET, UND-F, and UND-J2 datasets.


For age classification task, we have five different age groups based on the following age ranges: 18-28, 29-38, 39-48, 49-58, 59-68+. These classes have been selected according to the previous work about age classification from ear images \cite{yaman2018age}. According to the \cite{yaman2018age}, the changes in the ear can be observable between these age groups. These age classes include 419, 435, 316, 169, and 58 ear and profile face images. While the number of images for train, validation, and test is 1110, 144, and 143 respectively. Note that we have split data into train-validation-test sets with respect to the data distribution per class.

\begin{table}
\begin{center}
\begin{tabular}{cc|cc}
\hline
Model     & Data    & Age Acc. & Gender Acc. \\
\hline\hline
VGG-16    & Ear     &  60.97\%  & 97.56\%   \\
\hline
VGG-16    & Profile &  65.73\%  & 95.81\%   \\ 
\hline
ResNet-50 & Ear     &  60.97\%  & 98.00\%   \\ 
\hline
ResNet-50 & Profile &  62.37\%  & 94.05\%  \\ 
\hline
\end{tabular}
\end{center}
\caption{Unimodal age and gender classification results on FERET dataset~\cite{FERET_dataset}. }
\label{table:unimodal_age}
\end{table}

In all experiments, we have set learning rate to 0.0001. We have dropped learning rate by 0.1 in every 25 epochs. We have also used L2 regularization with 0.001 coefficient and center loss \cite{center_loss} with 0.1 coefficient as mentioned in Section \ref{Section CNN}. Moreover, to prevent overfitting, 
we have dropped neurons with 75\% probability value during training. On the other hand, we have not dropped them in test phase. While we have selected batch size as 32 for the unimodal experiments, we have chosen batch size as 16 during multimodal experiments due to memory constraints. 

\subsection{Unimodal Results}\label{Section Unimodal Results}



In this section, we present unimodal experiments, which are based on profile face and ear images separately. We have fine-tuned VGG-16 \cite{VGG} and ResNet-50 \cite{resnet}  CNN models for age and gender classification tasks with FERET dataset~\cite{FERET_dataset}. The obtained results are presented in Table \ref{table:unimodal_age}. In this table, first column contains used  CNN model, second column contains information about data, which can be ear image or profile face image and, finally, last two columns show test accuracies of age and gender classification.

According to Table \ref{table:unimodal_age}, we have achieved the best age classification result with VGG-16 model using profile face images with 65.73\% classification accuracy. While ResNet-50 performance on profile face images is slightly lower than VGG-16 performance, both models have achieved similar accuracy with ear images. 
The best gender classification result which is 98.00\%, is obtained with ear images using ResNet-50 CNN model and VGG-16 result is very close to this result. Although age classification performance is better with profile face images, ear images are found to be  more useful than profile face images in gender classification. All these results indicate that ear and profile face images contain useful cues for age and gender classification. 
However, age classification needs further investigation because of the relatively low performance compared to that of gender classification.

\subsection{Multimodal Results}\label{Section Multimodal and Multitask Results}


\begin{table}
\begin{center}
\begin{tabular}{cc|cc}
\hline
Model     & Fusion      &    Age Acc.     & Gender Acc. \\
\hline\hline
VGG-16    & A-1         &    61.83\%          &  92.33\%         \\
ResNet-50 & A-1         &    57.49\%          &  91.63\%         \\
\hline
VGG-16    & A-2         &    \textbf{67.59\%} &  \textbf{99.11\%}\\
ResNet-50 & A-2         &    62.71\%          &  \textbf{99.11\%}\\
\hline
VGG-16    & A-3         &    62.05\%          &  93.03\%         \\
ResNet-50 & A-3         &    58.53\%          &  92.33\%         \\
\hline
VGG-16    & B         &    67.28\% &  98.16\% \\
ResNet-50 & B         &    66.44\%          &  97.56\%         \\
\hline
VGG-16    & C-1       &    63.76\%          &  97.90\%                \\
ResNet-50 & C-1       &    63.06\%          &  98.00\%                \\
\hline
VGG-16    & C-2       &    63.06\%          &  97.90\%                \\
ResNet-50 & C-2       &    62.02\%          &  98.00\%                \\
\hline
VGG-16    & C-3       &    63.06\%          &  97.90\%                \\
ResNet-50 & C-3       &    62.02\%          &  98.00\%                \\
\hline
VGG-16    & C-4       &    63.76\%          &  97.90\%                \\
ResNet-50 & C-4       &    63.06\%          &  98.00\%                \\
\hline
VGG-16    & C-5       &    63.76\%          &  97.90\%                \\
ResNet-50 & C-5       &    63.06\%          &  98.00\%                \\
\hline
\end{tabular}
\end{center}
\caption{Age and gender classification results of the three different fusion methods that are explained in Section \ref{Section Fusion Methods}. In the fusion column, A, B, and C correspond to data, feature, and score fusion methods, respectively. In method A, A-1, A-2, and A-3 are channel fusion, spatial fusion, and intensity fusion methods, respectively. In C, C1, C2, C3, C4, and C5 represent different confidence score calculation methods that are presented in Table~\ref{table:score-based_fusion_formulas}. C-1 means the first formula of the Table~\ref{table:score-based_fusion_formulas} is used, C-2 means the second formula in the table is employed and etc.}
\label{table:multimodal}
\end{table}

In this section, we have presented multimodal and multitask age and gender classification experiments using VGG-16 and ResNet-50 CNN architectures. Table \ref{table:multimodal} shows age and gender classification results based on different fusion methods. First column contains name of the employed CNN model as in the unimodal experiments. Second column contains fusion methods that are data (A), feature (B), and score (C) fusion methods, respectively. In this column, \textit{A-1} indicates channel fusion, \textit{A-2} indicates spatial fusion, and \textit{A-3} indicates intensity fusion that are explained in Section~\ref{Section Image-based Fusion}. The \textit{B} is the feature fusion method that is also explained in Section \ref{Section Feature-based Fusion}. Lastly, for \textit{C}, which is the score-level fusion, we have used five different confidence score calculation methods. These methods are explained in Section~\ref{Section Score-based Fusion}. Next two columns present test accuracies of the age and gender classification tasks, respectively. 

\begin{table}
\begin{center}
\begin{tabular}{c|c|c|c}
\hline
Method/Model          & Dataset      & Data    & Accuracy \\
\hline\hline
\multicolumn{4}{c}{Gender Classification} \\
\hline
Distance+KNN \cite{gnanasivam2013gender}    & Internal     & Ear     &  90.42\%    \\
GoogLeNet \cite{yaman2018age}       & Internal     & Ear     &  94.00\%    \\
BoF+SVM \cite{zhang2011hierarchical}         & UND-F        & Ear     &  91.78\%    \\
HIS+SVM \cite{lei2013gender}         & UND-F        & Ear     &  92.94\%    \\
HIS+SVM \cite{lei2013gender}         & UND-J2       & Ear     &  91.92\%    \\
Gabor+Voting \cite{khorsandi2013gender}    & UND-J2       & Ear     &  89.49\%    \\
BoF-SVM \cite{zhang2011hierarchical}         & UND-F        & Profile &  95.43\%    \\
BoF-SVM \cite{zhang2011hierarchical}         & UND-F        & Multi   &  97.65\%    \\
Ours          & FERET        & Multi   &  \textbf{99.11\%}    \\
Ours          & UND-F        & Multi   &  \textbf{100\%}      \\
Ours          & UND-J2       & Multi   &  \textbf{99.79\%}    \\
\hline\hline
\multicolumn{4}{c}{Age Classification} \\
\hline
Yaman et al. \cite{yaman2018age}       & Internal     & Ear     &  52.00\%    \\
Yaman et al.* \cite{yaman2018age}     & FERET & Ear  & 58.53\% \\
Ours          & FERET        & Multi   &  \textbf{67.59\%}    \\
\hline
\end{tabular}
\end{center}
\caption{Comparison of the proposed methods with previous works. While first part contains gender classification results, second part presents age classification results. 
According to the results, we have achieved the state-of-the-art results for age and gender classification. 
We have implemented the proposed method in \cite{yaman2018age} and to have a fair comparison, we have tested it also on FERET dataset \cite{FERET_dataset}. The result of this method on FERET dataset is presented with * symbol.} 
\label{table:comparison}
\end{table}

According to the experimental results, the best age classification performance is achieved with VGG-16 model using \textit{A-2} fusion method which is the spatial fusion. However, the feature-based results, \textit{B}, are pretty close to the data fusion method and feature fusion with ResNet-50 is around 4\% better than all the other fusion types with ResNet-50 model. Generally, except spatial fusion method, other two data fusion methods are not as powerful as feature-based and score-based fusion strategies especially for gender classification. 

In gender classification experiments, we have achieved the best performance again with spatial fusion as in the age classification experiments. Both the VGG-16 and ResNet-50 CNN models achieve 99.11\% classification accuracy with spatial fusion. 

All these results indicate that the spatial fusion and feature-based fusion methods are effective multimodal fusion strategies for extracting soft biometric traits from profile face and ear images. 

\subsection{Comparison with State-of-the-art}\label{Section Analysis of the Results}

In Table \ref{table:comparison}, we have compared age and gender classification performance of our proposed method with previous works. We have achieved the state-of-the-art performance on both tasks, however, used dataset in this work is not the same as the ones used in previous work \cite{yaman2018age} for age classification task. Because of that, we have implemented and used the previous work on age classification \cite{yaman2018age}. We have followed the same strategy with them and presented the obtained result with * symbol in the table. Besides, we have conducted gender classification experiments on the UND-F and UND-J2 datasets \cite{UND_datasets}, which are benchmark datasets for gender classification from ear images. As a result, we have obtained the state-of-the-art results on both datasets with our spatial fusion method and feature-based fusion method. We have achieved 100\% accuracy on the UND-F dataset \cite{UND_datasets} and 99.79\% accuracy on the UND-J2 dataset \cite{UND_datasets}.  
For age classification, we have otained 67.59\% accuracy, with spatial fusion, that is around 9\% better than the accuracy achieved by the method presented in \cite{yaman2018age}. 

\section{Conclusion}\label{Section Conclusion}
In this paper, we presented different end-to-end multimodal deep learning frameworks for age and gender classification, which takes profile face and ear images as input. We employed domain adaptation in order to adapt the pretrained CNN models, namely VGG-16 and ResNet-50, to the ear domain. To learn more discriminative features, we combined center loss with softmax loss. 
We conducted extensive experiments on the UND-F, UND-J2, and FERET datasets for gender classification, and on the FERET dataset for age classification. We showed that by combining profile face and ear image, we can achieve very high accuracies. Spatial fusion and feature fusion methods have led to significant performance improvements. We achieved the state-of-the-art gender classification accuracies on the UND-F and UND-J2 datasets. In addition, we improved the age classification accuracy. To the best of our knowledge, this is the first study on age classification using ear images that is conducted on a publicly available dataset, FERET, in contrast to the previous work~\cite{yaman2018age}, in which authors used an internal dataset. Compared to the reported age classification accuracy in this previous work~\cite{yaman2018age}, we achieved around 9\% improvement for five class age group classification on FERET dataset. 

{\small
\bibliographystyle{ieee}

















}

\end{document}